\documentclass[runningheads]{llncs}

\usepackage{eccv} 

\usepackage{eccvabbrv}
\usepackage{adjustbox}
\usepackage{graphicx}
\usepackage{booktabs}
\usepackage{pifont}

\usepackage[accsupp]{axessibility}  
\newcommand{\xmark}{\ding{53}}%

\usepackage{hyperref}

\usepackage{orcidlink}

\begin{document}

\title{GraVITON: Graph based garment warping with attention guided inversion for Virtual-tryon} 

\titlerunning{GraVITON}


\authorrunning{}
\author{Sanhita Pathak, Vinay Kaushik, Brejesh Lall}

\institute{}
\maketitle

\begin{abstract}



Virtual try-on, a rapidly evolving field in computer vision, is transforming e-commerce by improving customer experiences through precise garment warping and seamless integration onto the human body. While existing methods such as TPS and flow address the garment warping but overlook the finer contextual details. In this paper, we introduce a novel graph based warping technique which emphasizes the value of context in garment flow. Our graph based warping module generates warped garment as well as a coarse person image, which is utilised by a simple refinement network to give a coarse virtual tryon image. The proposed work exploits latent diffusion model to generate the final tryon, treating garment transfer as an inpainting task. The diffusion model is conditioned with decoupled cross attention based inversion of visual and textual information. We introduce an occlusion aware warping constraint that generates dense warped garment, without any holes and occlusion. Our method, validated on VITON-HD and Dresscode datasets, showcases substantial state-of-the-art qualitative and quantitative results showing considerable improvement in garment warping, texture preservation, and overall realism.


\end{abstract}

\section{Introduction}
\label{sec:intro}

With the evolving shopping trends, ecommerce platforms have started catering to the customer needs keeping in sync with the emerging requirements. In the apparel industry this has come into view as the virtual tryon, which can provide a real inshop experience to the customers. The image based tryon methods\cite{bai2022single,He2022StyleBasedGA} have proven to be more practical when compared to the 3D\cite{lal2019multi} models which require modelling of the person for a realistic tryon synthesis which is quite labor-some.

To produce a perfect tryon result, the person and garment variability has to be prioritised while formulating the tryon pipeline. Although various studies have synthesized compelling results on the benchmarks\cite{Morelli2022DressCH,Choi2021VITONHDHV,Han2017VITONAI}, there still exist some paucity in terms of realism.

The tryon technique was first introduced by VITON\cite{Han2017VITONAI}, which used TPS warping for solving the problem of warping garments in virtual tryon. CPVTON\cite{wang2018toward} preserved texture better than VITON, but lacked perfect alignment, while the flow based approaches\cite{xie2023gpvton,lee2022hrviton,bai2022single} learnt robust structural alignment but lacked texture consistency. Other methods\cite{xie2021towards,morelli2023ladi} focused more on improving the generation by using various synthesis models such as GANs and recently diffusion\cite{morelli2023ladi}. 
Amid all the advancements in various stages of virtual tryon, there are still considerable gaps such as learning better garment warp, handling occlusion, pose transformations, generating consistent texture, etc. present, that leave a great scope of improvement.


The current methods\cite{bai2022single,xie2023gpvton} typically model the flow as result of correlations (utilising either a simple convolution network or feature correlation) between features across garment and reference images(pose,agnostic). These approaches mainly encode the point wise correspondence between an image feature pair(s) while neglecting the intra-relations among pixels within regions\cite{luo2022learning}. There's a need to capture discriminative features for region and shape representations.
Thus, decoupling the garment context from the warping procedure, and simultaneously transferring the region and shape prior of garment context to warping network can aid in learning an optimal garment warp. 

Motivated from AGFLOW\cite{luo2022learning}, which introduces iterative graph based flow estimation, we propose a solution to the aforementioned problem on warping by building a novel graph based garment warping module, which embeds context into learning garment warp onto the warping pipeline. The proposed Graph based flow warping module (GFW) learns to match features conditioned on garment context, and allows objects spatial neighbourhood to be well aggregated and thus largely decreases the uncertainty of ambiguous warping of garment.



Diffusion models\cite{Dhariwal2021DiffusionMB} currently stand as the top-performing models; when compared to the flow and TPS based counterparts\cite{bai2016exploiting,xie2023gpvton,wang2018toward}. However, maintaining texture consistency during warping poses a challenge. Recent approaches, exemplified by LaDI-VTON\cite{morelli2023ladi}, StableViton\cite{kim2023stableviton}, dci-vton\cite{gou2023taming}, CAT-DM\cite{zeng2023cat} address this challenge by leveraging textual and visual context for virtual try-on generation, treating it as a conditional image inpainting task. To achieve this, LaDI-VTON proposes inversion module, where image features are extracted from CLIP image encoder and mapped to new features by a trainable network and then concatenated with text features. StableVTON utilises a ControlNet model conditioned on straight garment incorporating a zero-conv cross attention block. CAT-DM initiates a reverse denoising process with an implicit distribution generated by a pre-trained GAN-based model, thereby reducing the sampling steps without
compromising generation quality. It can be seen as a way to have the ability to use image prompt, but the generated image is only partially faithful to the prompted image\cite{ye2023ip}.
In the cross-attention module of LaDI-VTON\cite{morelli2023ladi}, merging straight cloth features and text features into the cross-attention layer only accomplishes the alignment of image features to text features, and potentially misses some image-specific information and eventually leads to only coarse-grained controllable generation with the reference image.
This leads to texture transfer artefacts in some scenarios. For a better tryon inversion, we propose Decoupled Cross Attention adaptor(DCAA), which adds an additional cross-attention layer only for image features \cite{ye2023ip}.

Diffusion based approaches such as \cite{morelli2023ladi} utilise complete warped garment(without any holes/occlusion) as input, which is usually achieved by TPS or affine based transformations\cite{wang2018toward}. Flow based methods\cite{bai2022single,xie2023gpvton} generate dense flow for garment warping and have better warping than their TPS counterparts, while suffering from artefacts such as holes(self occlusion due to hands) present in ground truth warped garment images. There is a need to generate complete warped garments utilizing flow, for its optimal use in diffusion based generation pipelines. We devise one such way by introducing an occlusion aware warp loss(OWL). This loss excludes the warped garment learning for the occluded/masked garment section and results in a complete garment for tryon.

The contributions of our proposed work are as follows:
\begin{itemize}
    \item We introduce a Graph based flow warping module(GFW), that guides the flow warping by providing pixel neighbourhood context into source and reference. To the best of our understanding, we are the first to introduce graph based technique for garment warping.
    \item We propose Occlusion Aware warp Loss(OWL) to enable the complete warped garment learning in case of self-occlusion present in ground truth garments.
    \item We propose a Decoupled Cross Attention Adaptor (DCAA), enriching latent space inversion for a realistic tryon.
    \item Extensive experimentation and rigorous validation demonstrates that our method achieves state-of-the-art performance compared to existing prominent methods.
\end{itemize}

\section{Related Works}
\subsection{Virtual tryon}
Given a set of straight cloth and a person image, the goal of virtual tryon is to seamlessly warp the garment and overlay it onto the target person image. The initial work that introduced the garment warping and a generated complete person tryon was VITON\cite{Han2017VITONAI}. Other methods\cite{yang2020towards,He2022StyleBasedGA,wang2018toward,xie2023gpvton,Han2019ClothFlowAF} followed a similar two stage warping and generation pipeline which learnt TPS or affine transformation parameters for computing garment warp, while the flow based methods\cite{He2022StyleBasedGA,Dosovitskiy_2015_ICCV} were introduced for garment warp following a similar two stage pipeline that changed the warping scenario for tryon. Although the texture preservation for TPS warping is superior to that of flow, flow still gains on the garment alignment with the changing human pose. In order to achieve the realism in the final tryon, it is crucial to formulate a robust deformation module. This is usually achieved by the deformation of control points with an energy function (radial basis function) in TPS based pipelines (Thin Plate Spline)\cite{wang2018toward}, and by computing per pixel appearance flow map followed by target view synthesis in flow based pipelines. 

The flow based warping learns dense pixel correspondence\cite{bai2022single} when compared to the TPS based methods, where the sparse distance between the control points plays a crucial role when the points are fit using the transformation function\cite{Han2017VITONAI}.
Both methods\cite{wang2018toward,Han2019ClothFlowAF} estimate a global deformation and hence fail to estimate the local deformations successfully for various body parts.
Other methods have focused on addressing the garment alignment\cite{lee2022hrviton} effectively.

\subsection{Graph neural networks in flow}

Optical flow is the task of estimating dense per-pixel correspondence between images. GMFlow\cite{xu2022gmflow} introduced vision transformers for computing optical flow, but its heavy computational dependencies made it less diversely applicable. AGFlow\cite{luo2022learning} exploited the scene/context information, utilising graph convolutional networks, and incorporated it in the matching procedure to robustly compute optical flow. 
GPVTON\cite{xie2023gpvton} tried to address the local deformations by applying a part wise flow based deformation, where the garment is disintegrated and deformed separately into three regions, one for each upper body part but it is not able to jointly optimise the local and global deformations. Another work KGI\cite{Li_2023_ICCV} utilised graph to predict the garment pose points guided by human pose which inpainted the predicted region using human segmentation. The method failed to achieve the precision in tryon alignment due to sparse guiding points to guide the dense pixel warping for garment texture unlike in flow methods. Hence, motivated by AGFlow\cite{luo2022learning}, in this work we have shown that GCNs can help the garment warping by focusing on the pixel level deformations establishing a dense correlation that helps in preserving the local details post deformation, which is ideally faced by all the flow based garment warping methods. 

\subsection{Diffusion Models}
Diffusion models marked research has become a foundational area in the field of image synthesis \cite{Dhariwal2021DiffusionMB} because of its high quality image generation.
Tasks such as image-to-image translation \cite{saharia2022palette}, image editing \cite{avrahami2022blended}, text-to-image synthesis \cite{gu2022vector}, and inpainting \cite{lugmayr2022repaint,nichol2021glide} have seen significant progress due to their realistic generation results. \cite{jiang2022text2human} concentrated on creating full-body images by sampling from a trained texture-aware codebook, given human position and textual descriptions of clothing shapes and textures.
Furthermore, in order to address the problem of pose-guided human prediction, \cite{bhunia2023person} created a texture diffusion block that was conditioned by multi-scale texture patterns from the encoded source image. Adding to the tryon generation features, \cite{baldrati2023multimodal} introduced using the model pose, the garment sketch, and a textual description of the garment to condition the tryon generation process. Building on these methods and to improve the texture generation in person tryon, LaDI-VTON\cite{morelli2023ladi} utilised a textual inversion component, enabling mapping of garment visual features to the CLIP token embedding space. This process generates a set of pseudo-word token embeddings, effectively conditioning the generation process. DCI-vton\cite{gou2023taming} leverages a warping module to combine the warped clothes with clothes-agnostic person
image, and add noise as the input of diffusion model to guide the diffusion model’s generation effectively. Other methods on diffusion such as StableVITON\cite{kim2023stableviton} and CAT-DM\cite{zeng2023cat} utilises a ControlNet model conditioned on straight garment for tryon.

\begin{figure}
    \centering
    \includegraphics[width=\linewidth, height=8.3cm]{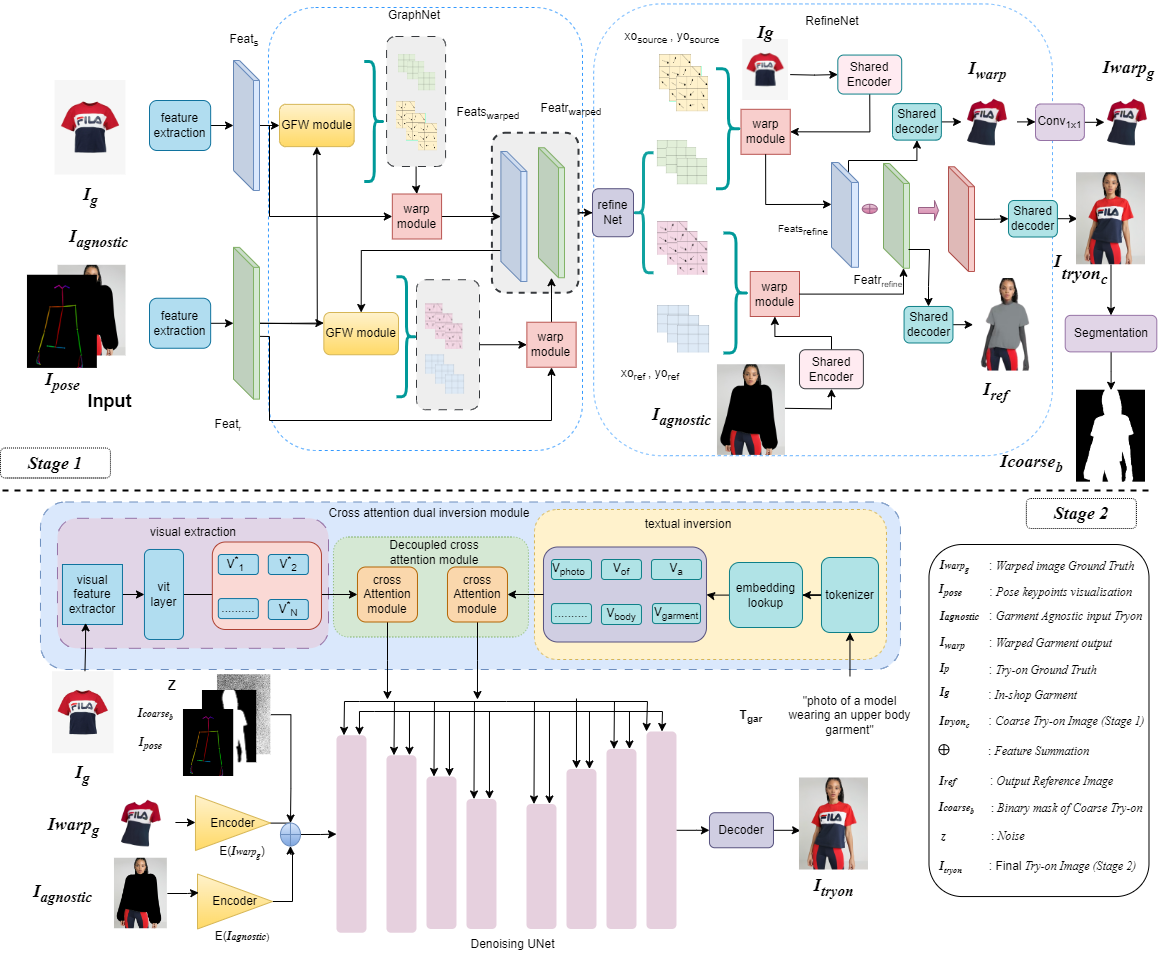}
    \caption{Architecture Diagram of GRAVITON. The top module utilizes GCNs for generating warped cloth and coarse tryon image. These outputs are processed to condition the Stable Diffusion model. The inversion model efficiently computes Cross-Modal attention to improve texture and structural consistency, generating the final tryon image.}
    \label{fig:fig1}
    \vspace{-0.9cm}
\end{figure}

\section{Proposed Approach}
Our proposed model employs a two-stage pipeline. In the first stage, it performs warping, while the second stage generates the final tryon result. The first stage of the pipeline comprises of a graph based warping module, which is followed by a refinement module. The inputs to this stage are in the form of source garment($I_{g}$) and reference input concatenated as (reference pose ($I_{pose}$), agnostic image ($I_{agnostic}$)). The warping stage computes the dense flow with graph correlation volume giving warped garment($I_{warp_g}$) and coarse tryon($I_{tryon_c}$) as an output. Furthermore, it is guided by a loss constraint $Loss_{owl}$ producing the complete garment($I_{warp_g}$) without any textural irregularities.
The second stage of the pipeline synthesizes the final tryon results using the diffusion model in an inpainting approach. The inputs to this stage are the person segmentation mask ($I_{coarse_b}$) computed from the coarse tryon ($I_{tryon_c}$), and warped output from stage one ($I_{warp_g}$), human pose keypoints($I_{pose}$), agnostic input ($I_{agnostic}$) and noise ($I_{z}$) as an input. The diffusion process is conditioned with the attention based inversion between textual data ($T_{gar}$), coarse garment($I_{tryon_c}$) and source cloth for texture ($I_{g}$). The calculated decoupled attention conditions the latent space to generate final tryon($I_{tryon}$). 


\subsection{Graph based coarse tryon}
The coarse tryon stage caters to the generation of warped garment along with coarse tryon that is further used in final tryon generation in stage 2.
The input to the first stage, is source garment($I_{g}$), reference pose ($I_{pose}$) and agnostic image ($I_{agnostic}$). The network employs a features extraction module in form of basic convolution layers with N=3, N being the number of conv layers and a stride 2.

\subsubsection{GraphNet}
The features extracted for both source($Feat_{s}$) and reference($Feat_{r}$) features further act as an input to GFW module that returns the offsets and attention for source and feature warping. The reference features to the GFW module act as the context for coarse flow calculation.


Features $Feat_{s}$, $Feat_{r}$ are utilised for dense deformable flow prediction.
This obtained deformable flow is calculated for \textbf{m=6} 2D flows. The average flow captures the possible degree of $\Delta$ flow with the given value of m. The average offset addition provides the final dense flow.

\begin{equation}
f_o = (x_{o},y_{o}) = \dfrac{\sum_{k=1,m} (\delta x_{m}, \delta y_{m})} {\sum_{k=1,m} 1}
    \label{eq:one}
\end{equation}

The dense flow offsets $(x_{o},y_{o})$ along with the computed attention maps are utilised by the warping module to warp $Feat_{s}$ feature to compute source warped feature $Feat_{s_{warped}}$. 

Similarly, the source warped feature $Feat_{s_{warped}}$ and reference feature $Feat_{r}$ 
are fed to the GFW module to compute reference warped feature $Feat_{r_{warped}}$. 

\subsubsection{RefineNet}
The \textbf{refineNet} module computes attention between generated person and warped garment along with a finer offset for better warping and coarse tryon\cite{bai2022single}. 


The concatenated source and reference warped features $Feat_{s_{warped}}$ and $Feat_{r_{warped}}$ are fed as the input to refineNet module to compute final offsets $x_{o_{source}},y_{o_{source}}$ and $x_{o_{ref}},y_{o_{ref}}$ which are the final warping directives for source garment ($I_{g}$) and reference input ($I_{agnostic}$) respectively, giving the  source refined features $Feat_{s_{refine}}$ and reference refined features $Feat_{r_{refine}}$. Both source and reference refined features are summed and sent to a shared decoder to compute the warped output coarse tryon image($I_{tryon_c}$). Similarly, the source refined feature $Feat_{s_{refine}}$ is fed to the shared decoder to compute generated warped garment image ($I_{warp}$), which is further refined by being passed through a 1x1 convolution layer to compute $I_{{warp}_{g}}$. 

Both $I_{warp_g}$ and $I_{tryon_c}$ are utilised by  generation stage (stage 2).


\subsubsection{Graph based Flow Warping module(GFW)}
The warping in graph network projects a highly connected space providing a dense pixel context utilising graph's adjacency property.
\begin{figure}
    \centering
    \includegraphics[width=\linewidth, height=6cm]{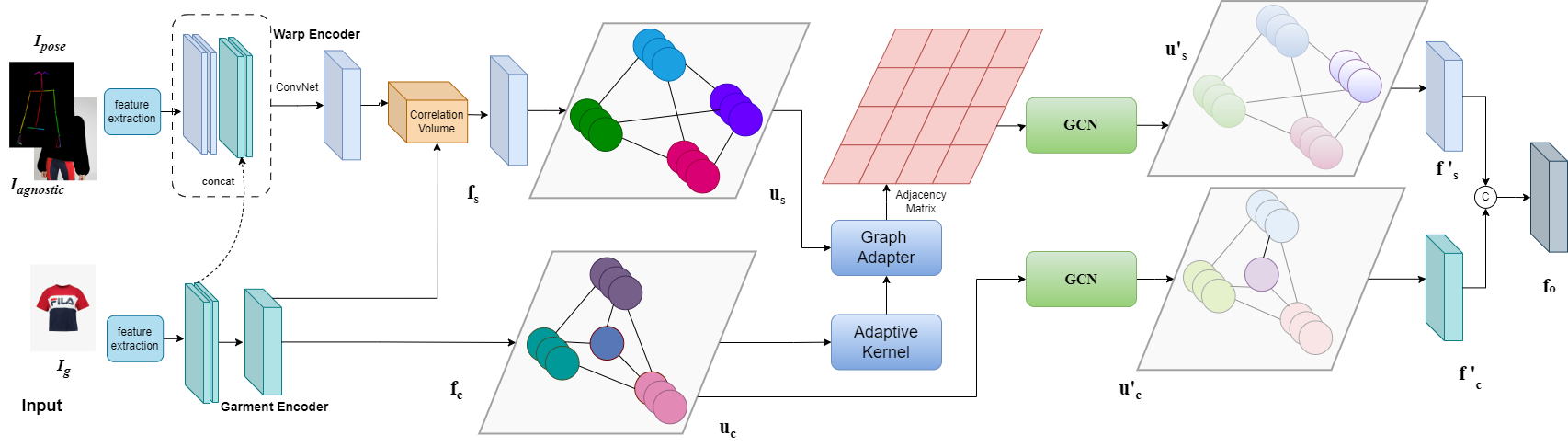}
    \caption{Architecture Diagram of Graph based Flow  Warping Module (GFW). The module utilizes GCNs for generating warped cloth and coarse tryon image.}
    \label{fig:fig1g}
\end{figure}
The source and reference features $Feat_{s}$, $Feat_{r}$ are utilized to construct a 4D correlation volume capturing the statistical similarity between the two. The resulting value is sent to four convolutions to capture motion feature $f_{s}$ and the reference feature ($Feat_{r}$) is fed to the garment encoder network to compute context feature $\textbf f_{c}$ as shown in Figure \ref{fig:fig1g}. Both features are utilised to perform a holistic warp reasoning by computing offsets $\textbf f_{o} = (x_{o},y_{o})$.


The graph based module in stage1 consists of nodes($N$) and edges($E$) formulated in a directed graph as $G$=($N$,$E$).
The node embeddings are mapped to the graph space using a simple projection function, $u = \textbf{{P}}_{f\rightarrow u} (f)$, where \textbf{u} denotes the nodes in graph space, \textbf{P} is the projection function and \textbf{f} depicts the feature space. 
We define the nodes mapped into context (garment) feature $\textbf f_{c}$ and warp feature $\textbf f_{s}$ encoded as,
${\mathbf U}_c$ = ($u_{c}^{1}$,$u_{c}^{2}$,.....,$u_{c}^{n}$) and ${\mathbf U}_s$ = ($u_{s}^{1}$,$u_{s}^{2}$,.....,$u_{s}^{n}$), where ${\mathbf U}_c$ is context nodes for garment warping while 
${\mathbf U}_s$ is the normalized feature correlation between features of the source (garment) and reference (pose, agnostic) in the graph space.




The process of node creation for both the source and context entails the computation of the adjacency matrix, which measures the similarity between all nodes denoted as ${\mathbf U}_c$ and ${\mathbf U}_s$. To facilitate adaptive graph learning, we employ $\cal L()$ as a graph learner, comprising of a two-layer convolutional network with ReLU activation. The first layer focuses on channel-wise learning for ${\mathbf U}_s$, while the second layer introduces node-wise interaction learning, resulting in a refined node representation for the source denoted as $\hat{\mathbf U}_s^{(t)}$.

\begin{equation}
	\breve{\mathbf A_{s}} = {\cal L}({\mathbf U}_s ; \Theta({\mathbf U}_c));    \hat{\mathbf U}_s^{(t)} = {\cal F}_{\tt AG}({\mathbf u}_s, \breve{\mathbf A})^{(t)})
 \label{eq:adj}
\end{equation}

\begin{equation}
	\hat{\mathbf U}_c^{(t)} = {\cal F}_{\tt Graph}({\mathbf U}_c, {\mathbf A})^{(t)}, {\tt where}~ {\mathbf A} = {\mathbf U}_c^{\mathsf T}{\mathbf U}_c,
 \label{eq:context}
\end{equation}

The final adjacency matrix for context and warp nodes is formulated in equation \ref{eq:adj} giving the modified nodes for the source, with $\Theta ()$ signifying a parameter learner and ${\cal F}_{\tt AG}$ is adaptive graph learning function for warping.
The context nodes are computed as in equation \ref{eq:context} where ${\cal F}_{\tt Graph}$, is graph learning function in the Graph Adapter block defining the warping context.


The projection function \textbf{P} preserves the spatial details during the first(initial) conversion to the graph space, and utilising this, the modified nodes are projected back from graph to feature space using the projection function \textbf{P} as shown in equation \ref{eq:eight} and equation \ref{eq:nine}, giving $\hat{\mathbf f}_c$ garment (context) and source warp feature $\hat{\mathbf f}_s$. 

\begin{equation}
	\hat{\mathbf f}_c = {\mathbf f}_c + h {\cal P}_{{\mathrm v}\rightarrow {\mathrm f}}({\mathbf u}_c),
 \label{eq:eight}
\end{equation}
where, $h$ denotes a learnable parameter that is initialized as $0$ and gradually performs a weighted sum. Similarly, the source warp feature $\hat{\mathbf f}_s$ is produced by

\begin{equation}
	\hat{\mathbf f}_s = {\mathbf f}_s + l {\cal P}_{{\mathrm s}\rightarrow {\mathrm f}}({\mathbf u}_s).
 \label{eq:nine}
\end{equation}

where, $l$ denotes a learnable parameter The resultant features are then concatenated to give the resulting offsets on the original grid from source image.

\begin{equation}
    \mathbf{f_{o}} = (1+ {F_{ch}}(\hat{\mathbf f}_s)) * concat(\hat{\mathbf f}_c, \hat{\mathbf f}_s)
\end{equation}

where, ${F_{ch}}$ signifies the channel attention.
\par $\mathbf{f_{o}} =  (x_{o}^g , y_{o}^g) = (\delta x , \delta y)$  from equation \ref{eq:one}.

\subsubsection{Loss: Occlusion Aware warp Loss(OWL)}
In flow based methods while learning the garment warping the dense deformation leads to the learning of self occlusion due to human poses and limb positions. As a result the resulting warped garment may not have the correct semantic structure post warping and results in a sub-optimal tryon generation due to inferior warped input. To handle this problem our work proposes an occlusion aware loss that learns the warping of the garment even for the occluded garment spaces.
The mask generated for $I_{gt}^{warp}$ consists of binary values for the warped siloutte, these pixels when multiplied with the L1 distance between $I_{gt}^{warp}$ and $I_{warp_g}$ enhance the occluded parts of garment resulting in the learning of complete garment.

The input to the second stage for generation of final tryon utilises this loss only for warped garments from the flow stage to give clean input to diffusion.

Input- $I_{gt}^{warp}$ and $I_{warp_g}$

\begin{equation}
    Mask_{gt} = Binary\_threshold(I_{gt}^{warp})
\end{equation}

\begin{equation}
L_{owl} = \dfrac{\sum_{i=1,M} \sum_{j=i,N} Mask_{gt}^{ij} * (I_{warp_{gt}}^{ij} - I_{warp_g}^{ij})} {\sum_{i=1,M} \sum_{j=i,N}Mask_{gt}^{ij}   }
    \label{eq:owl}
\end{equation}

\par The overall loss is presented as where ${L}_{style},{L}_{prec},{L}_{L1}$ are style , perceptual and L1 losses.
\begin{equation}
\mathcal{L} = (\lambda_{L1} \mathcal{L}_{L1} + \lambda_{prec} \mathcal{L}_{prec} + \lambda_{style} \mathcal{L}_{style} +\lambda_{owl}  \mathcal{L}_{owl})
    \label{eq:loss_s1}
\end{equation}

\subsection{Cross Modal Attention for Inversion}
Stage one outputs $I_{warp_g}$ , $I_{tryon_c}$ act as inputs to the tryon generation stage.
The $I_{warp_g}$ image for warped garment is used as an input to the diffusion model. Ladivton\cite{morelli2023ladi} constrains the generation of tryon for a fixed pose, which is due to the parsing inputs required by the diffusion model which limits the diverse pose generative capacity of the model. To solve this, we utilise the coarse tryon output $I_{tryon_{c}}$ from stage 1 to compute all the preprocessing inputs at stage 2 including person agnostic$I_{agnostic}$, binary person segmentation mask$I_{coarse_{b}}$, as well as pose keypoints$I_{pose}$. These preprocessed inputs go into the diffusion model for training.

\subsubsection{Diffusion model}:
The model consists of an encoder \textbf{E} and decoder \textbf{D} block encapsulated within an autoencoder \textbf{A}. Also, a time conditioned U-net is used with a denoising parameter $\epsilon$. 
The diffusion encoder takes in the warped garment and person agnostic processed by a shared encoder ${E}$ giving the warped encoded garment ${E}({I}_{Warp_{g}})$ and  encoded person agnostic ${E}({I}_{agnostic})$. The additional inputs: pose $I_{pose}$, mask $I_{coarse_{b}}$ and noise $z$ are resized to the encoded spatial size and concatenated. 


The resulting inputs to the network are combined as: \\ $\beta = [Z; I_{coarse_{b}}; I_{pose}; {E}({I}_{Warp_{g}}); {E}({I}_{agnostic})]$ and used for latent learning. The stable diffusion model is used as an in-painting approach as in ladivton \cite{morelli2023ladi}. To make the learning more accurate the latent space is conditioned with the textual embedding and garment tryon image $I_{tryon}$ for texture embedding utilising our DCAA module.

As the tryon aims to transfer the given warped garment to the person, it can be dealt as an inpainting task inspired by \cite{morelli2023ladi}. Our proposed framework focuses to inpaint the masked area, but instead of being guided by a TPS based warped garment, our diffusion model is guided by the warped garment computed from stage 1.
A CLIP encoder is employed for textual inversion which takes textual data${T_{gar}}$ as an input. Similarly, input straight cloth $I_g$ is fed to a pretrained variational encoder, and the features are fed to a ViT layer to compute texture feature for the same. The texture features from image are represented in CLIP token embedding space, similar to \cite{morelli2023ladi}. The token embeddings from the textual data acts as a textual prompt that guides the garment texture positioning. To enhance this, we introduce a decoupled attention adaptor to condition the Denoising UNet giving realistic tryon results. 


\subsubsection{Decoupled Cross Attention Adaptor (DCAA)}
Although methods such as LaDI-VTON\cite{morelli2023ladi} use inversion to enhance the diffusion process, still it lacks to learn optimal results and can generate tryon with erroneous texture details. This happens because the image features are not effectively embedded in the pretrained model, as they simply feed the concatenated features to the cross-attention layers. By inducing the features in such a way to diffusion models, it fails to capture the fine-grained features from image prompt. To solve this problem, we propose the Decoupled Cross Attention Adaptor. 
Hence, embedding the image features using newly added cross-attention layers is an effective strategy which improves the feature understanding and embedding in overall inversion process. 
The textual features obtained from the CLIP embedding ${x}_{t}$ are fed into the cross attention layer along with the query features z, given by latent. Hence, the cross-attention equation is given as, 

\begin{equation}
\begin{split}
\mathbf{z}'=\text{Attention}(\mathbf{\alpha},\mathbf{\beta},\mathbf{\gamma}) = \text{Softmax}(\frac{\mathbf{\alpha}\mathbf{\beta}^{\top}}{\sqrt{d}})\mathbf{\gamma},
\\
\end{split}
\label{eq:decouple_img}
\end{equation}

where, $\alpha = zW_{\alpha}$, $\beta = x_{i}W_{\beta}$
and $\gamma = x_{i}W_\gamma$ are the query, key, and values matrices from the text features and $W_{\beta}, W_{\gamma}$.
are the corresponding weight matrices. In DCAA, the cross attention layers for text features and garment features are separate. We add a new cross attention layer, for each cross attention layer in the original UNet model to insert garment features. Given the garment features $g_{i}$, the output of new cross attention $\mathbf{z}''$ is computed as follows:

\begin{equation}
    \begin{split}
    \mathbf{z}''=\text{Attention}(\mathbf{\alpha},\mathbf{\beta}',\mathbf{\gamma}') = \text{Softmax}(\frac{\mathbf{\alpha}(\mathbf{\beta}')^{\top}}{\sqrt{d}})\mathbf{\gamma}',\\
    \end{split}
    \label{eq:decouple_txt}
\end{equation}

where, $\alpha = zW_{\alpha}$, $\beta' = g_{i}W'_{\beta}$
and $\gamma' = g_{i}W'_\gamma$ are the query, key, and values matrices from the image features and $W'_{\beta}, W'_{\gamma}$.
are the corresponding weight matrices.

We use the same query for image cross-attention as for text cross-attention. Consequently, we only need to add two paramemters $W'_{\beta}$ and $W'_{\gamma}$ for each cross-attention layer. In order to speed up the convergence, $W'_{\beta}$ and $W'_{\gamma}$ are initialized from $W_{\beta}$ and $W_{\gamma}$. 

Combining both the equations, \ref{eq:decouple_img} and \ref{eq:decouple_txt} we get the final cross attention equation as below,

\begin{equation}
\begin{split}
\mathbf{z}^{new}=\text{Softmax}(\frac{\mathbf{\alpha}\mathbf{\beta}^{\top}}{\sqrt{d}})\mathbf{\gamma}+\text{Softmax}(\frac{\mathbf{\alpha}(\mathbf{\beta}')^{\top}}{\sqrt{d}})\mathbf{\gamma}'\\
 \text{where}\ \mathbf{\alpha}=\mathbf{z}\mathbf{W}_\alpha, \mathbf{\beta}=\boldsymbol{x}_{t}\mathbf{W}_\beta, \mathbf{\gamma}=\boldsymbol{x}_{t}\mathbf{W}_\gamma,
 \mathbf{\beta}'=\boldsymbol{x}_{i}\mathbf{W}'_\beta, \mathbf{\gamma}'=\boldsymbol{x}_{i}\mathbf{W}'_\gamma
\end{split}
\end{equation}

Here, $W'_{k}$ and $W'_{v}$ are trainable weights while others are frozen.


\subsection{Training losses}
\textbf{Loss}: The diffusion noise learns from the loss function as defined in ladi-vton \cite{morelli2023ladi} for stage 2 training.
For stage 1 training the loss is derived from equation\ref{eq:loss_s1}

\begin{figure}
    \centering
    \adjustbox{width=\linewidth}{\begin{tabular}{c  c | c  c}
\multicolumn{2}{c}{LaDI-VTON(TPS)} & \multicolumn{2}{c}{Proposed(Graph Flow)} \\

        \includegraphics[width=0.25\linewidth, height=25mm]{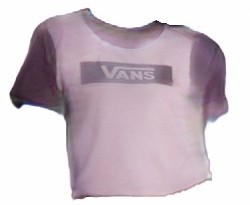} &
        \includegraphics[width=0.25\linewidth, height=25mm]{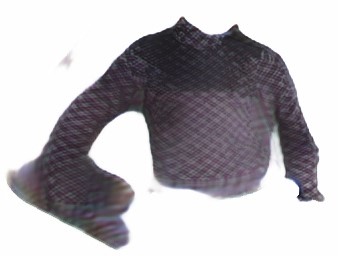} &
        \includegraphics[width=0.25\linewidth, height=25mm]{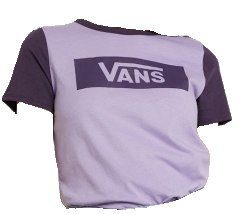} &
        \includegraphics[width=0.25\linewidth, height=25mm]{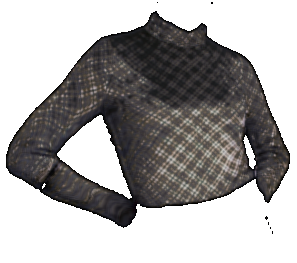}
    \end{tabular}
    





}
    \caption{Qualitative comparison between the warping methodologies of proposed method with LaDI-VTON}
    \label{fig:fig2}
\end{figure}

\section{Experiments}
\subsection{Dataset}
The experiments were conducted on VITON-HD and Dresscode datasets. VITON-HD is a high resolution dataset with resolution of 1024x768. The train set consists of 11,647 train pairs and 2,032 test pairs. DressCode is composed of 48,392/5,400 training/testing pairs of front-view full-body person and garment from different categories (i.e., upper, lower, dresses). The model is trained for both datasets in a paired setting on upper body garments and tested on both paired and unpaired setting. The same garment tryon is tested on the model as it is wearing in paired. While, a different garment tryon is tested on the model in an unpaired setting.

\begin{figure}
    \centering
    \adjustbox{width=\linewidth}{\centering
\begin{tabular}{c c c c c c} 
Person & Garment & VITON-HD\cite{Choi2021VITONHDHV} & HR-VITON\cite{lee2022hrviton} & LaDI-VTON\cite{morelli2023ladi} & Proposed \\
        \includegraphics[width=0.25\linewidth, height=32mm]
        {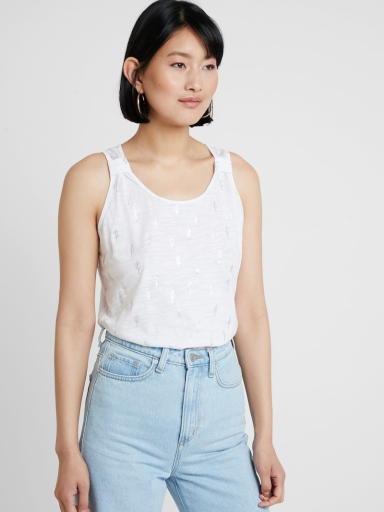} &
        \includegraphics[width=0.25\linewidth, height=32mm]
        {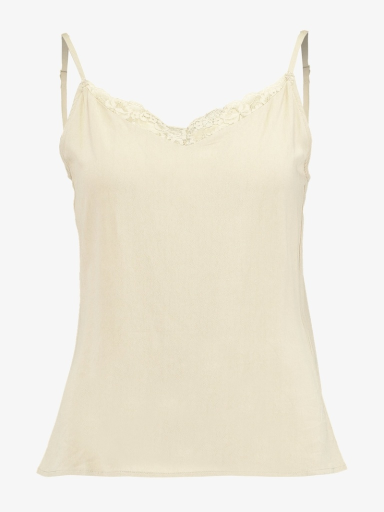} &
        \includegraphics[width=0.25\linewidth, height=32mm]{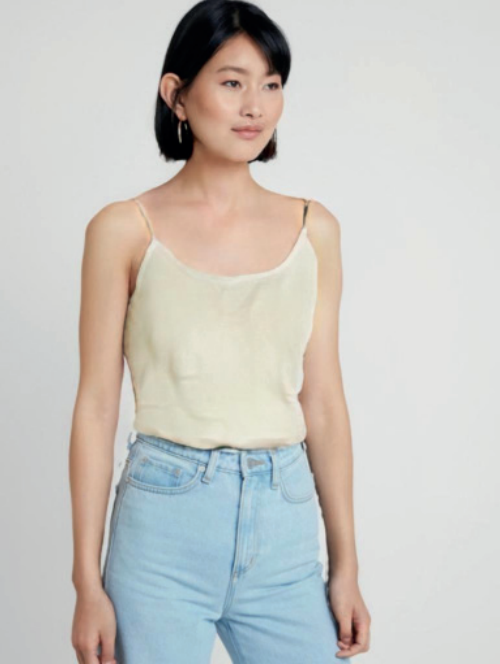} &
        \includegraphics[width=0.25\linewidth, height=32mm]{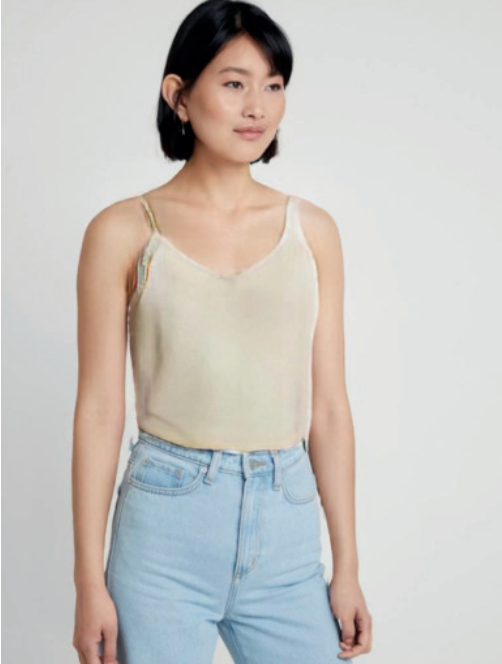} &
        \includegraphics[width=0.25\linewidth, height=32mm]{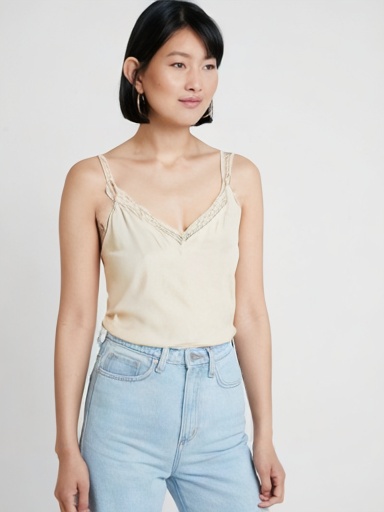} &
        \includegraphics[width=0.25\linewidth, height=32mm]{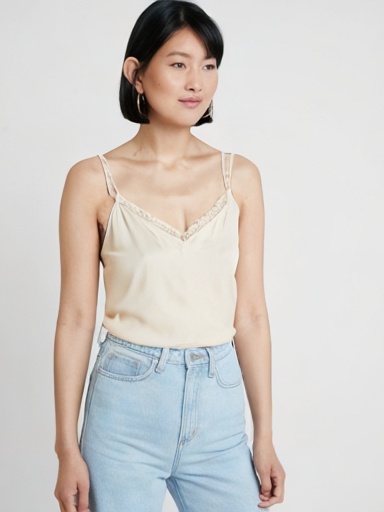}
          \\
        \includegraphics[width=0.25\linewidth, height=32mm]
        {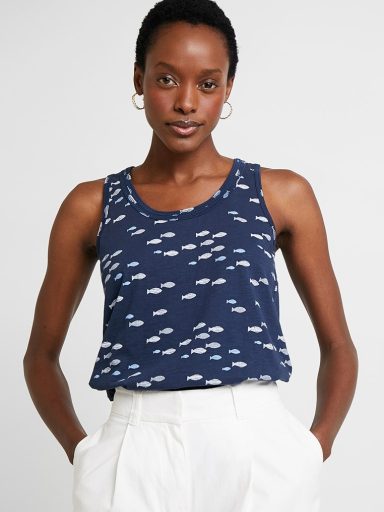} &
        \includegraphics[width=0.25\linewidth, height=32mm]
        {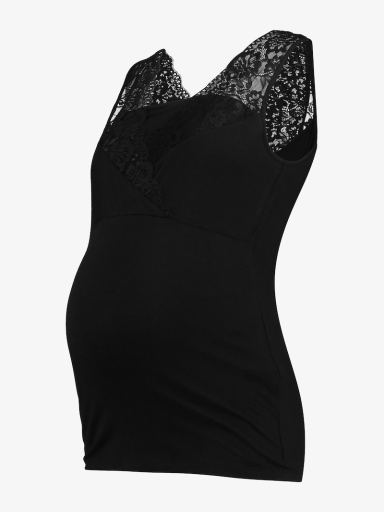} &
        \includegraphics[width=0.25\linewidth, height=32mm]{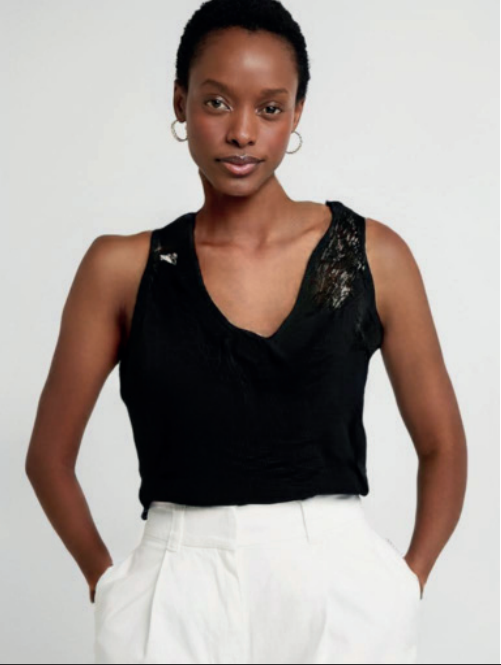} &
        \includegraphics[width=0.25\linewidth, height=32mm]{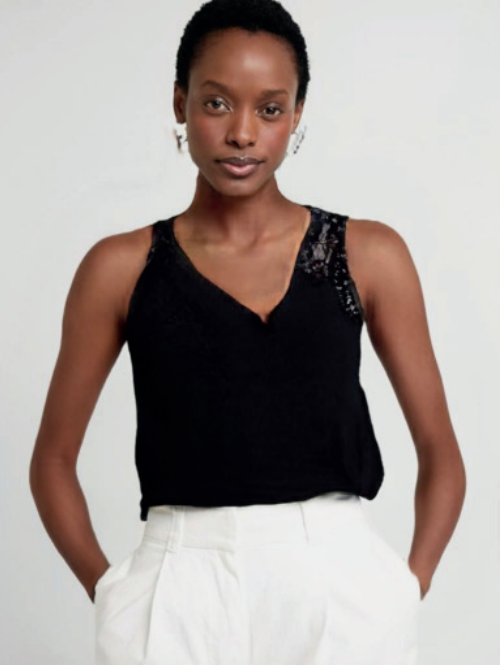} &
        \includegraphics[width=0.25\linewidth, height=32mm]{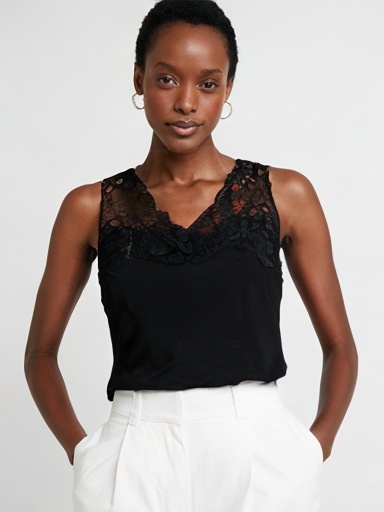} &
        \includegraphics[width=0.25\linewidth, height=32mm]{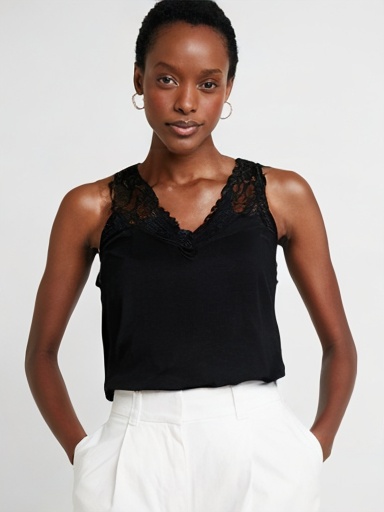}
         \\
    \end{tabular}
}
    \caption{Qualitative results generated by Proposed method in comparison with VITON-HD, HR-VTON, LaDI-VTON}
    \label{fig:fig3}
\end{figure}
\subsection{Implementation Details and Training}
The model is trained in two stages successively. The graph based warping stage is trained first by computing the dense flow and the resulting warped garment and coarse tryon are the input to second stage.
The experiments were conducted on pytorch on one V100 GPU. Our model was trained for 200 epochs, for a batch of 6 with a learning rate of 0.000035. Weights for the loss functions are $\lambda_{L1}=1, \lambda_{prec}=1, \lambda_{style}=100$.

For stage two training, the inputs derived from stage 1 are employed and the preprocessed coarse tryon is used as a pose guiding feature. The tryon is trained jointly with decoupled attention prompt.
We used AdamW as training optimiser with $\beta1$= 0.9, $\beta2$  = 0.999 and weight decay equal to 1e-2.
The attention decoupled adaptor is trained with the diffusion model.

To evaluate our model, we use metrics such as LPIPS and SSIM to assess coherence with ground-truth images. For realism, we employ FID and KID metrics in both paired and unpaired settings. We use torch-metrics for LPIPS and SSIM, and \cite{parmar2022aliased} for FID and KID scores. This comprehensive evaluation framework enables rigorous assessment of our model's fidelity and realism. 

\begin{figure}
    \centering
    \adjustbox{width=11cm,height=4cm}{\centering
\begin{tabular}{c c c c c c}
\centering
Input Cloth & Person & Graph Tryon & Proposed(w/o attention) & LadiVTON\cite{morelli2023ladi}  & Proposed \\
        \includegraphics[width=0.25\linewidth, height=32mm]{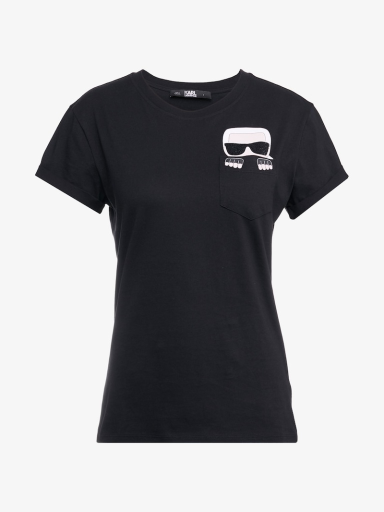} &
        \includegraphics[width=0.25\linewidth, height=32mm]{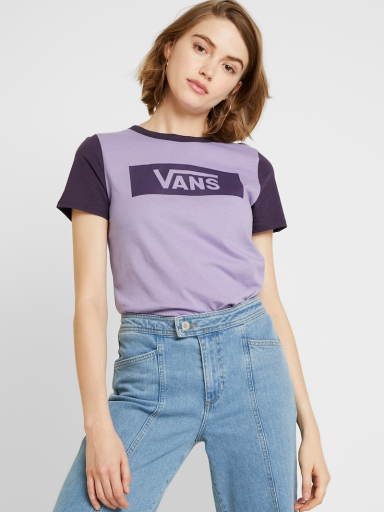} &
        \includegraphics[width=0.25\linewidth, height=32mm]{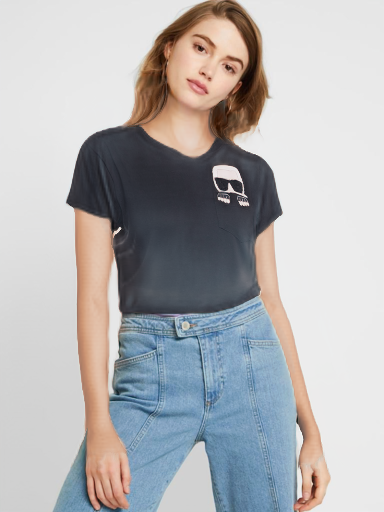} &
        \includegraphics[width=0.25\linewidth, height=32mm]{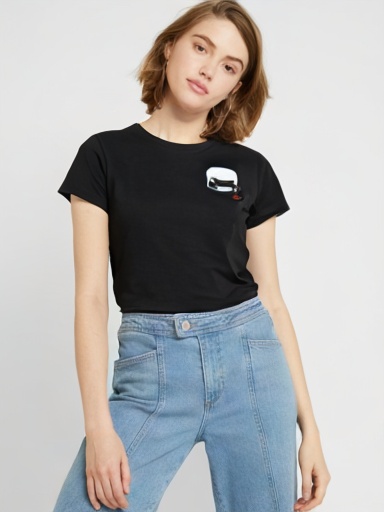} &
        \includegraphics[width=0.25\linewidth, height=32mm]{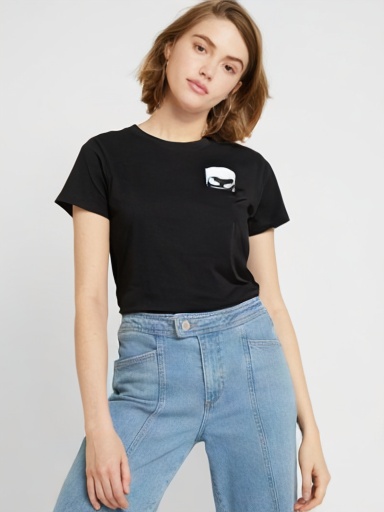} &
        \includegraphics[width=0.25\linewidth, height=32mm]{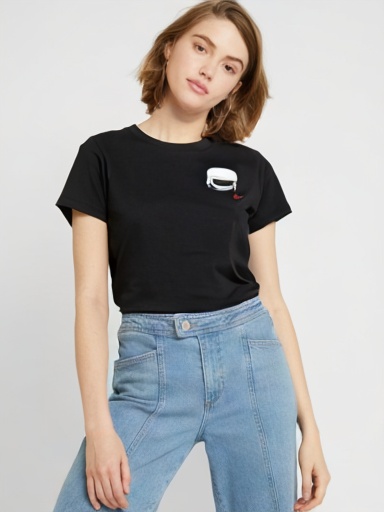} \\

        \includegraphics[width=0.25\linewidth, height=32mm]{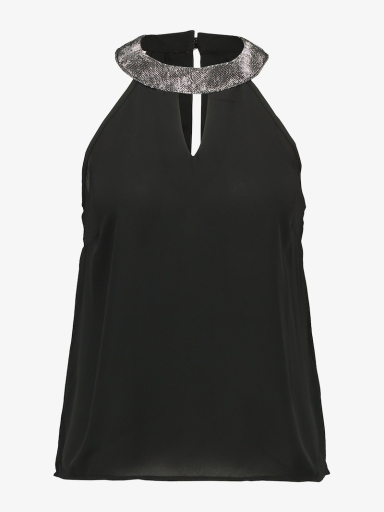} &
        \includegraphics[width=0.25\linewidth, height=32mm]{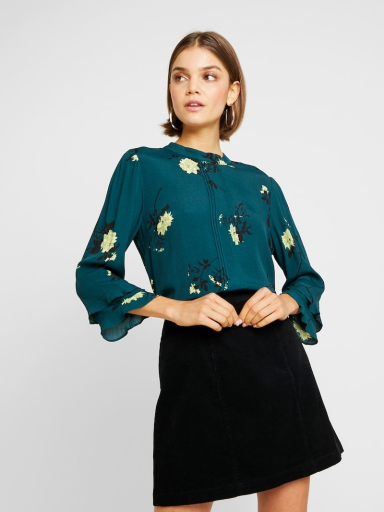} &
        \includegraphics[width=0.25\linewidth, height=32mm]{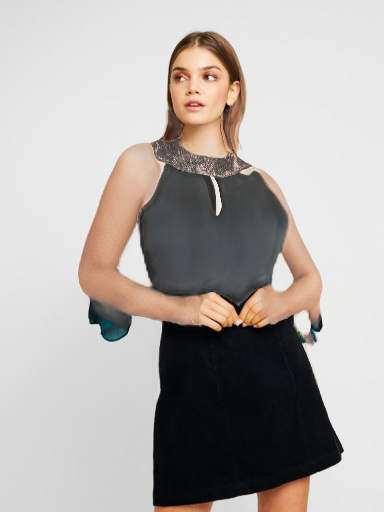} &
        \includegraphics[width=0.25\linewidth, height=32mm]{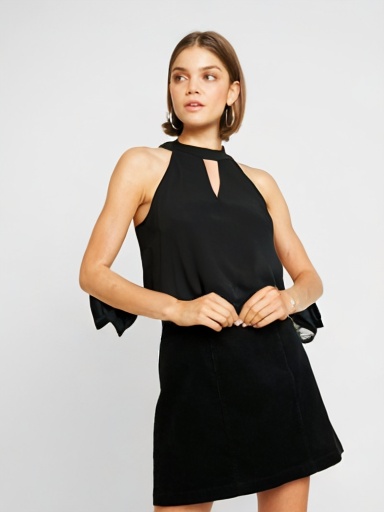} &
        \includegraphics[width=0.25\linewidth, height=32mm]{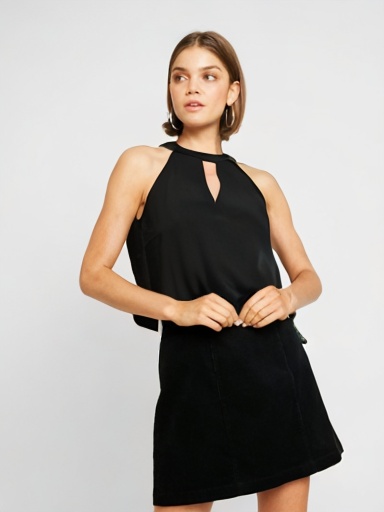} &
        \includegraphics[width=0.25\linewidth, height=32mm]{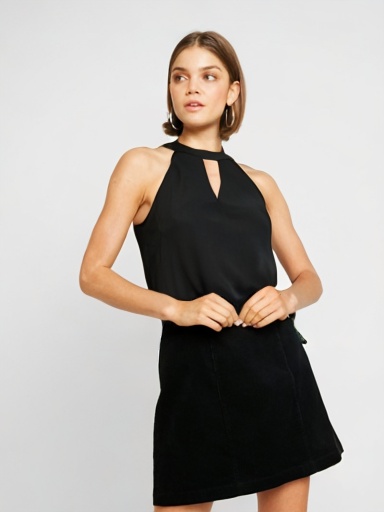} \\
        \includegraphics[width=0.25\linewidth, height=32mm]{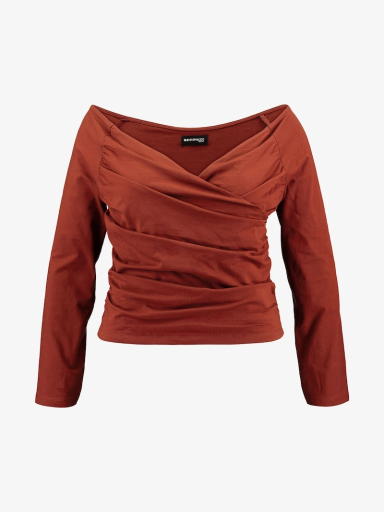} &
        \includegraphics[width=0.25\linewidth, height=32mm]{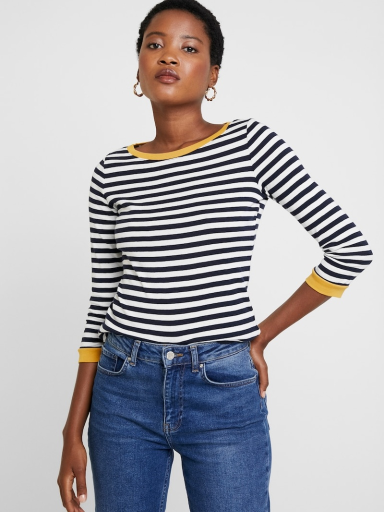} &
        \includegraphics[width=0.25\linewidth, height=32mm]{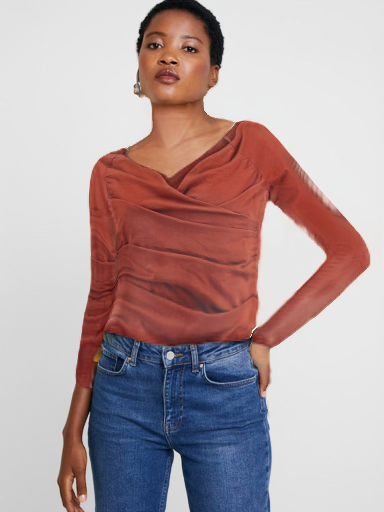} &
         \includegraphics[width=0.25\linewidth, height=32mm]{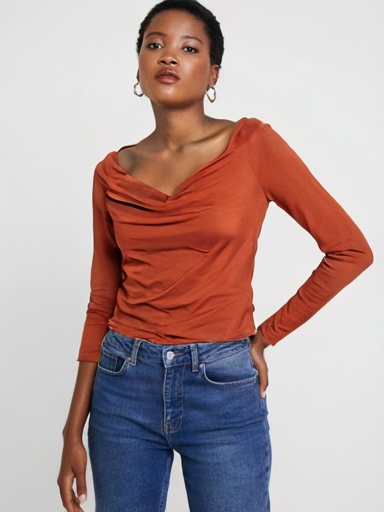} &
        \includegraphics[width=0.25\linewidth, height=32mm]{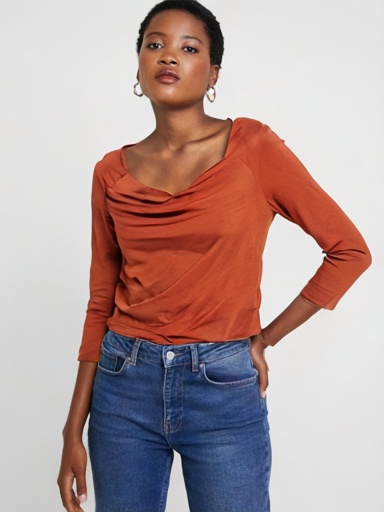} &
        \includegraphics[width=0.25\linewidth, height=32mm]{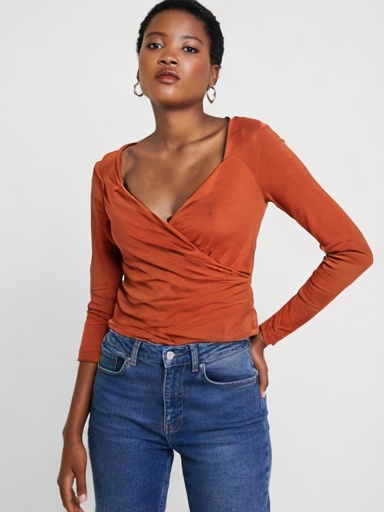} \\
        \includegraphics[width=0.25\linewidth, height=32mm]{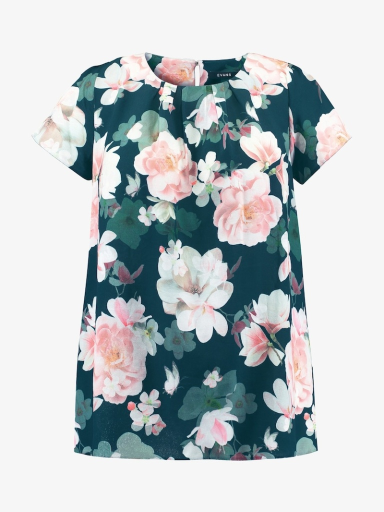} &
        \includegraphics[width=0.25\linewidth, height=32mm]{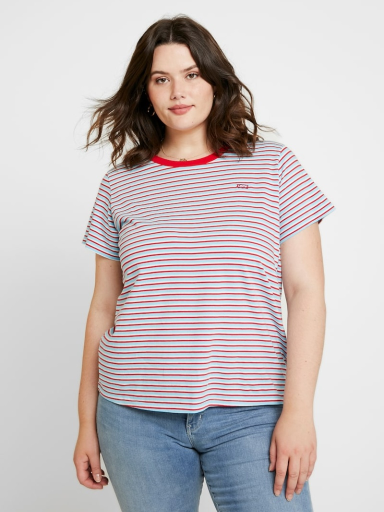} &
        \includegraphics[width=0.25\linewidth, height=32mm]{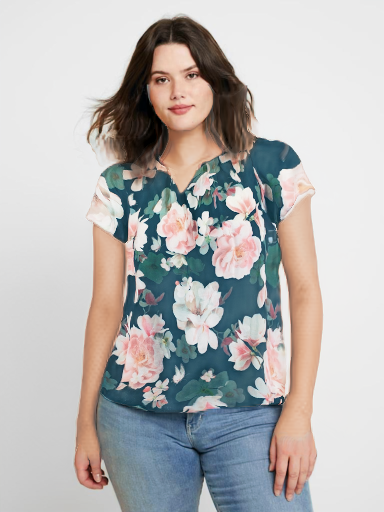} &
        \includegraphics[width=0.25\linewidth, height=32mm]{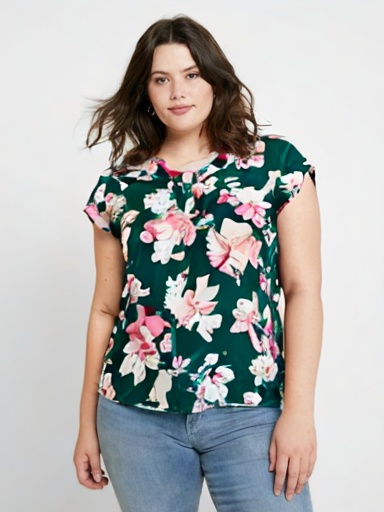} &
        \includegraphics[width=0.25\linewidth, height=32mm]{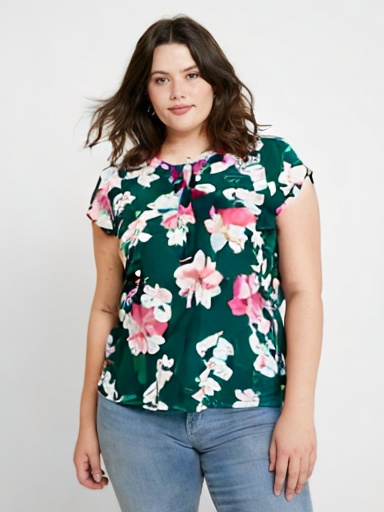} &
        \includegraphics[width=0.25\linewidth, height=32mm]{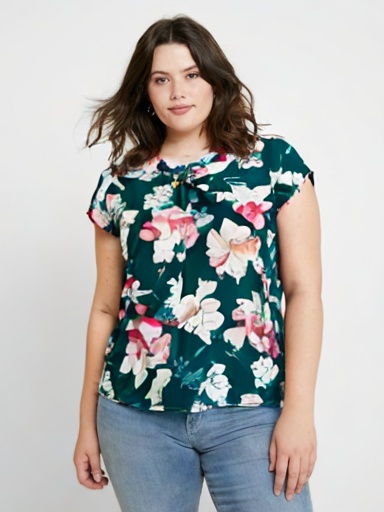} \\
    \end{tabular}}
    \caption{Tryon reconstruction Qualitative results showing the successive visual enhancement in results and comparative details with LaDI-VTON }
    \label{fig:fig4}
    \vspace{-0.5cm}
\end{figure} 

\begin{table*}[t]
\centering
\begin{tabular}{l|l|l|lll}
                    & SSIM & FID  & KID  \\ \hline \hline
Graph based tryon &   0.88   &  7.87   &   1.8        \\ \hline
Flow based tryon  &  0.864    &  8.49   &    2.1       \\ \hline
Diffusion with flow &   0.87  &  7.61  & 1.2       \\ \hline
\textbf{Diffusion with graph}     &   \textbf{0.89}   &      \textbf{6.57}    &  \textbf{1.06}  \\ \hline
\end{tabular}
\caption{Quantitative comparison between proposed method and incremental modules on VITON-HD dataset for paired setting}
\label{tab:tab1}

\end{table*}

\subsection{Qualitative Results}
To qualitatively assess our findings, we present sample images generated by our model alongside those generated by competing methods in Figure \ref{fig:fig3}. Notably, our approach demonstrates the capability to produce highly realistic images while preserving the intricate textures and details of the original in-shop garments. Furthermore, our model accurately maintains the physical characteristics of the target models. VITON-HD warps garment correctly but fails to estimate local warping and thus misses fine details in garments. HR-VITON improves upon VITON-HD and learns better warp and texture preservation. LaDI-VTON introduces diffusion to Virtual Tryon and brings considerable improvements to texture details present in the garment. Our proposed approach improves the texture consistency, utilising the decoupled attention based inversion module and learns a better garment warp, utilising a Graph based deformable flow estimation framework to warp garment, instead of TPS\cite{wang2018toward}. 

In Figure \ref{fig:fig4}, we systematically compare the outputs from each stage and module of our proposed methodology with those derived from LaDi-VTON, a diffusion based generative method in the domain of virtual try-on synthesis. Commencing with the Graph Tryon output, representing the coarse try-on image generated via the initial stage of our framework, we observe a foundational representation of the garment's warp onto the target model. Leveraging graph-based flow warping, this stage facilitates an initial alignment of the garment onto the target body, thereby establishing a baseline for subsequent refinement. This stage is lightweight and though it lacks rich texture information present in other stages as shown in Figure \ref{fig:fig4}, but it has correct global garment warp which aids in the subsequent diffusion stage. In Stage 2, the diffusion model is employed for refining the warped garment over the agnostic image. We present two variants: one devoid of attention in inversion module and another incorporating attention-based inversion. The latter imbues finer details and coherence into the resultant tryon output. 

Contrasting these outcomes with those of LaDi-VTON, which lacks graph-based garment warping and attention mechanisms within its inversion module, we discern notable disparities in the fidelity and realism of the virtual try-on results. Through meticulous visual inspection in Figure \ref{fig:fig4}, we discern improvements such as texture preservation, micro-texture retention in green top(last row), spatial coherence in black dress(second row), and consistent boundary warp(third row). While our proposed methodology showcases advancements in coherence and fidelity through the utilization of graph-based warping and attention-driven refinement, LaDi-VTON demonstrates remarkable proficiency in texture generation and realism; however, it does not reach comparable levels of precision and detail rendition as our proposed method.

\begin{figure}
    \centering
    \includegraphics[width=\linewidth]{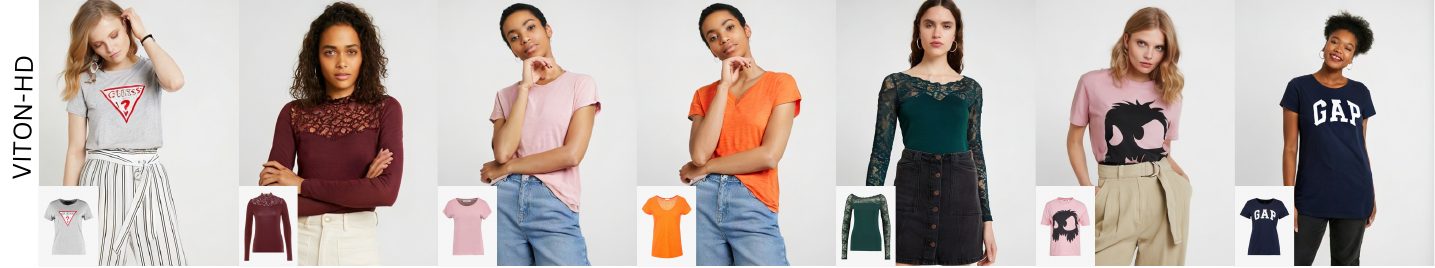}
    \caption{Qualitative results of our proposed methodology on VITON-HD Dataset depicting pose, hair, sleeve length and texture variations.}
    \label{fig:qual5.pdf}
\end{figure}

\begin{figure}
    \centering
    \includegraphics[width=\linewidth]{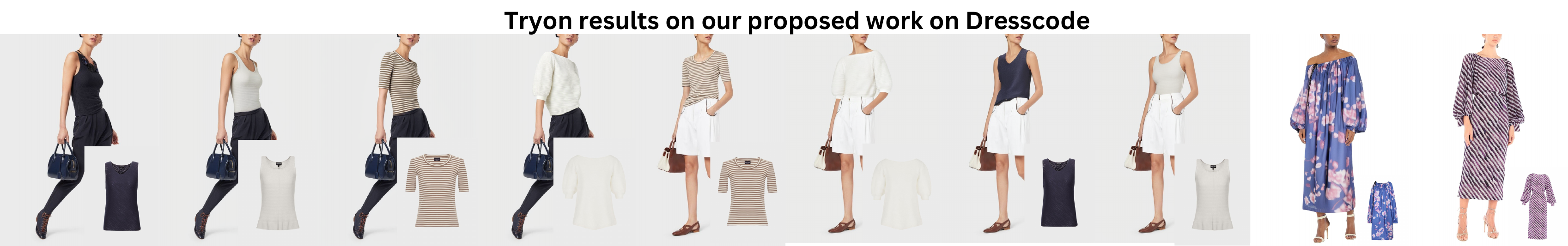}
    \caption{Qualitative results of our proposed methodology on Dresscode Dataset depicting pose, sleeve length, upper/dress and texture variations.}
    \label{fig:qual6.pdf}
    \vspace{-1cm}
\end{figure}

Figure\ref{fig:qual5.pdf} depicts the garment tryon in an unpaired setting for VITON-HD dataset with texture variations, sleeve lengths, pose and hair. The generated images provide visual effectiveness of our method to handle self occlusion due to complex arm positions as can be seen in four images from the left. The proposed method also generates realistic garment textures retaining the fine details of text and symbols in the images.Our method also handles tryon with complex poses preserving the texture, sleeve-length and hair fidelity.
Figure\ref{fig:qual6.pdf} shows realistic tryon generation for Dresscode dataset. Our work generates realistic tryon for both upper garments and dresses in unpaired setting. The results preserve texture, sleeve length and are agnostic to pose variations.


\subsection{Quantitative results}
We describe the robustness and correctness of our proposed approach by conducting extensive experiments and ablation on Dresscode and VITON-HD datasets.
Table \ref{tab:tab1} demonstrates that the affect of introduction of graph for garment warping and coarse try-on prediction improves the accuracy of try-on module significantly when compared with the flow based traditional counterparts.
It also describes the improvement in final try-on after utilising our diffusion model for target person generation.
As we see, combination of Graph and Diffusion achieves the best result quantitatively.

\begin{table*}[t]
\begin{minipage}{0.48\textwidth}
\centering
\begin{tabular}{l|ll}
\hline
OWL & SSIM & FID \\ \hline
\xmark  &   0.87   &  6.71   \\ \hline
\checkmark &  0.89   &   6.57  \\ \hline
\end{tabular}
\vspace{0.2cm}
\caption{Affect of occlusion aware loss }
\label{tab:tab2}
\end{minipage}%
\hfill
\begin{minipage}{0.48\textwidth}
\centering
\begin{tabular}{l|l|l}
                              & SSIM & FID \\ \hline
Attention in inversion         & 0.891   &  6.57    \\ \hline
No attention in inversion &   0.878   &  6.93   \\ \hline
\end{tabular}
\caption{Affect of attention in inversion module VITON-HD}
    \label{tab:tab4}
\end{minipage}
\vspace{-1cm}
\end{table*}



Table \ref{tab:tab3} describes how various flow modules aid in warping input garment. The iterative flow which was motivated from RAFT\cite{luo2022learning} is unable to learn optimal warp, as the flow being learnt is an intermediate component of our network. While, RAFT\cite{luo2022learning} being a supervised framework introduced a flow consistency constraint which utilises ground truth flow that aids in learning of the iterative flow, we learn flow as an intermediate component in self-supervised manner. We also see that introduction of deformable flow\cite{bai2022single} to our graph based flow estimation framework drastically improves learning of warped garment. This enhancement can be attributed to the fusion of features warped using multiple flows, resulting in the creation of a single optimized try-on. Consequently, while individual warped features may exhibit slight discrepancies, the fusion process aggregates the most favorable attributes from all features to generate an optimal try-on output.

\begin{table}[]
\centering
\begin{tabular}{l|l|l|}
                    & SSIM & FID  \\ \hline \hline
Iterative flow &  0.84    &  8.6        \\ \hline
Single stage flow  &    0.85   & 8.2       \\ \hline
Deformable flow &   0.89   &  6.5       \\ \hline
\end{tabular}
\caption{Flow method comparison on VITON-HD dataset}
\label{tab:tab3}
\vspace{-1cm}
\end{table}


The introduction of decoupled cross attention between text embedding and garment texture feature embedding improves the consistency of texture learnt in final tryon. This can be seen as improvement in FID and SSIM scores in table \ref{tab:tab4}.

The Occlusion Aware warp loss(OWL) aims to mask incorrect holes present in ground truth warped garment. This ensures that the network doesnot learn to predict warp in those regions. Table \ref{tab:tab2} shows improvement in SSIM and FID due to introduction of this loss in the pipeline of stage 1.



We analyzed the impact of each individual component on the performance of the model. As shown in Table \ref{tab:tab5}, incorporating graph-based flow estimation led to a notable improvement in SSIM scores, indicating enhanced spatial coherence and perceptual quality in the generated images. Similarly, the integration of diffusion mechanisms in the generation process resulted in significantly lower FID scores, demonstrating improved fidelity and realism in the synthesized outputs.

Furthermore, the inclusion of attention mechanisms within the inversion module led to substantial gains in both SSIM and FID metrics, highlighting the importance of selective feature extraction and reconstruction in enhancing image quality and content preservation. Additionally, incorporating occlusion-aware masking loss functions contributed to further improvements in both SSIM and FID scores, indicating better handling of garment artifacts and garment boundary preservation.

Most notably, our comprehensive approach, combining all key components yielded the most impressive results. As depicted in Table \ref{tab:vitonhd_all}, our proposed approach achieved the highest SSIM and lowest FID scores among all prominent tryon methods on VITON-HD dataset, demonstrating the synergistic effects of our holistic technique.


\begin{table}[t]
    \caption{Quantitative results on the VITON-HD dataset~\cite{Choi2021VITONHDHV}. The * marker indicates results reported in previous works.\vspace{-0.2cm}}
    \label{tab:vitonhd_all}
    \setlength{\tabcolsep}{.3em}
    \resizebox{\linewidth}{!}{
    \begin{tabular}{lc cc cccc}
    \toprule
    \textbf{Model} & & \textbf{LPIPS} $\downarrow$ & \textbf{SSIM} $\uparrow$ & \textbf{FID$_\text{p}$} $\downarrow$ & \textbf{KID$_\text{p}$} $\downarrow$ & \textbf{FID$_\text{u}$} $\downarrow$ & \textbf{KID$_\text{u}$} $\downarrow$ \\
    \midrule
    CP-VTON*~\cite{wang2018toward} & & - & 0.791 & - & - & 30.25 & 40.12 \\
    ACGPN*~\cite{yang2020towards} & & - & 0.858 & - & - & 14.43 & 5.87 \\
    \midrule
    VITON-HD~\cite{Choi2021VITONHDHV} & & 0.116 & 0.863 & 11.01 & 3.71 & 12.96 & 4.09\\
    HR-VITON~\cite{lee2022hrviton} & & 0.097 & 0.878 & 10.88 & 4.48 & 13.06 & 4.72\\
    LaDI-VTON~\cite{morelli2023ladi} & & 0.091 & 0.876 & 6.66 & 1.08 & 9.41 & 1.60 \\
     \midrule
    \textbf{Proposed} & & \textbf{0.088} & \textbf{0.891}& \textbf{6.57} & \textbf{1.06} & \textbf{9.20} & \textbf{1.46} \\
    \bottomrule
    \end{tabular}
    }
\vspace{-.35cm}
\end{table}

\begin{table}[]
\centering    
\begin{tabular}{l|l|l|l|l|l}
\textbf{Graph} & \textbf{Diffusion} & \textbf{Attention Inversion} & \textbf{OWL} & \textbf{SSIM} & \textbf{FID} \\ \hline \hline
\xmark     & \xmark         & \xmark                    & \xmark   & 0.86 & 8.49   \\ \hline
\checkmark     & \xmark         & \xmark                    & \xmark   & 0.88 &  7.87   \\ \hline
\checkmark     & \checkmark         & \xmark                    & \xmark   & 0.88 & 7.23   \\ \hline
\checkmark     & \checkmark         & \checkmark                    & \xmark   & 0.87 & 6.63 \\ \hline
\checkmark     & \checkmark         & \checkmark                    & \checkmark & \textbf{0.89} &  \textbf{ 6.57 } \\ 
\end{tabular}

\caption{Quantitative Ablation of our proposed modules on VITON-HD dataset}
    \label{tab:tab5}
\end{table}




\section{Conclusion}
Our paper introduces novel solutions to enhance virtual try-on technology, addressing critical challenges in garment warping and generation. By incorporating the Graph-based Flow Warping module (GFW), we achieve more accurate context reasoning, significantly reducing uncertainty in garment transfer. Our Occlusion Aware Warp Loss (OWL) effectively handles self-occlusion, ensuring finer garment learning and seamless integration onto the human body. Additionally, the Decoupled Cross-Attention Mechanism (DCAA) enriches latent space information, leading to more realistic try-on synthesis. Empirical validation on benchmark datasets demonstrates substantial improvements in garment warping, texture preservation, and overall realism compared to existing methods. Our contributions significantly improve the seminal work done by previous approaches by proposing novel graph-based framework for garment warping and introducing novel pose try-on synthesis using diffusion models.

%
%
\bibliographystyle{splncs04}
\bibliography{main}
\end{document}